\pdfoutput=1
\documentclass[letterpaper, 10pt, conference]{ieeeconf}
\IEEEoverridecommandlockouts
\usepackage{cite}
\usepackage{amsmath,amssymb,amsfonts}
\usepackage{graphicx}
\usepackage{times}
\usepackage[table]{xcolor}
\usepackage{booktabs}
\usepackage{array}
\usepackage{pifont}
\usepackage{stfloats}
\usepackage{tikz}
\usetikzlibrary{positioning,arrows.meta,shapes.geometric,patterns,calc}

\definecolor{clrSotaA}{RGB}{255,127,14}
\definecolor{clrSotaB}{RGB}{0,160,135}
\definecolor{clrOurs}{RGB}{31,119,180}
\definecolor{clrRef}{RGB}{214,39,40}

\newcommand{\cmark}{\ding{51}}
\newcommand{\xmark}{\ding{55}}

\newcommand{\ControlHz}{20}
\newcommand{\ObsDistM}{0.23}
\newcommand{\ExplTriggerConf}{0.6}
\newcommand{\ExplTriggerN}{15}
\newcommand{\RecoveryConf}{0.8}
\newcommand{\RecoveryK}{5}
\newcommand{\BlindGraspDist}{0.02}
\newcommand{\CapsuleStop}{0.05}
\newcommand{\CapsuleSlow}{0.15}
\newcommand{\LinkRadius}{0.035}
\newcommand{\ExplTimeoutS}{10.0}

\newcommand{\NumTrialsPerCond}{50}

\newcommand{\EIHSuccess}{52\%}
\newcommand{\EIHPoseErr}{18.3}
\newcommand{\EIHCycle}{48}

\newcommand{\FixedOverallSuccess}{64\%}
\newcommand{\FixedRecovery}{8.2$\pm$2.1}
\newcommand{\FixedPoseErr}{12.7}
\newcommand{\FixedCycle}{45}

\newcommand{\StaticSuccess}{70\%}
\newcommand{\StaticRecovery}{6.8$\pm$1.9}
\newcommand{\StaticPoseErr}{11.4}
\newcommand{\StaticCycle}{46}

\newcommand{\ReactiveSuccess}{76\%}
\newcommand{\ReactiveRecovery}{4.5$\pm$1.4}
\newcommand{\ReactivePoseErr}{10.4}
\newcommand{\ReactiveCycle}{47}

\newcommand{\OursOverallSuccess}{94\%}
\newcommand{\OursRecoveryMeanS}{1.8}
\newcommand{\OursRecoveryStdS}{0.6}
\newcommand{\OursPoseErr}{6.6}
\newcommand{\OursCycle}{49}

\newcommand{\AblSFSuccess}{76\%}
\newcommand{\AblSFPoseErr}{14.5}

\newcommand{\AblTPSuccess}{80\%}
\newcommand{\AblTPPoseErr}{8.7}

\newcommand{\AblNoExplSuccess}{72\%}
\newcommand{\AblNoExplPoseErr}{7.1}

\newcommand{\AblSMSuccess}{86\%}
\newcommand{\AblSMPoseErr}{7.8}

\newcommand{\CalPertThirtyFullSucc}{86\%}
\newcommand{\CalPertThirtyNoSFSucc}{30\%}

\begin{document}

\title{\LARGE \bf
Active Observation for Robust Visual Servoing in Robotic Manipulation%
\begin{tikzpicture}[remember picture, overlay]
  \node[anchor=north, yshift=-6.5mm, text width=0.90\paperwidth, align=center, font=\normalfont\footnotesize\color{black!75}] at (current page.north) {%
    Author's accepted version of an article published in \emph{2026 IEEE 22nd International Conference on Automation Science and Engineering (CASE)}, pp.~3026--3031.\\
    Final published version available at DOI:~10.1109/CASE69030.2026.11704293.%
  };
  \node[anchor=south, yshift=5mm, text width=0.88\paperwidth, align=center, font=\normalfont\scriptsize\color{black!75}] at (current page.south) {%
    \copyright~2026 IEEE. Personal use of this material is permitted. Permission from IEEE must be obtained for all other uses, in any current or future media, including reprinting/republishing this material for advertising or promotional purposes, creating new collective works, for resale or redistribution to servers or lists, or reuse of any copyrighted component of this work in other works.%
  };
\end{tikzpicture}%
}

\author{Zhilun Hu$^{1}$,
Yuqi Cheng$^{1}$,
Yuxiang Tan$^{1}$,
Wenpan Li$^{1}$,
and Weiming Shen$^{1}$,~\IEEEmembership{Fellow,~IEEE}%
\thanks{This work was supported in part by the National Natural Science Foundation of China under Grant No.~52188102. (\textit{Corresponding author: Weiming Shen.})}%
\thanks{$^{1}$Zhilun Hu, Yuqi Cheng, Yuxiang Tan, Wenpan Li, and Weiming Shen are with the National Center of Technology Innovation for Intelligent Design and Numerical Control, Huazhong University of Science and Technology, Wuhan 430074, China (e-mail: wshen@ieee.org).}%
}

\maketitle
\thispagestyle{empty}
\pagestyle{empty}

\begin{abstract}
Robust visual servoing during robotic grasping remains challenging, as approach-phase occlusion frequently disrupts closed-loop control. To address this, we present an active observation framework in which one arm of an existing dual-arm platform carries an RGB-D camera as an active observer while the other executes grasping. The observer repositions to maintain or recover a shared target--gripper view within the feasible shared-view workspace.
Occlusion during the final approach is particularly acute: fixed eye-to-hand lacks an active recovery mechanism once the target is hidden, while eye-in-hand can suffer from self-occlusion and sensitivity to hand--eye calibration.
The observer first uses an active two-phase viewpoint strategy to align high-precision RGB-D axes with task-relevant error components. If visibility later degrades, an occlusion-triggered exploration module pauses the worker arm and scores candidate viewpoints by observation quality to recover a shared target--gripper view. Because both objects are detected in a single RGB-D image, the resulting \emph{Relative Servo Error} avoids differencing two independently transformed target and gripper estimates.
We evaluate the system on a dual 6-DoF Piper platform using an open-vocabulary detector with text prompts, without task-specific fine-tuning or demonstrations.
Across \NumTrialsPerCond{} randomized trials per object, the system achieves \OursOverallSuccess{} grasp success versus \ReactiveSuccess{} for a reactive viewpoint baseline and \FixedOverallSuccess{} for a fixed camera. On triggered severe-occlusion trials, recovery takes \OursRecoveryMeanS{}$\pm$\OursRecoveryStdS{}\,s on average.
\end{abstract}

\section{Introduction}

Robotic manipulation increasingly relies on visual feedback for accurate and flexible interaction in unstructured environments. Visual servoing closes the perception--action loop for such tasks: the robot continuously observes the scene, estimates the discrepancy between the current and desired configuration, and updates its motion online, compensating for perception errors, kinematic uncertainty, and environmental disturbances. It has thus become a fundamental tool for visually guided robotic manipulation~\cite{chaumette2006visual,chaumette2007visual,xu2025vsreview}.

Despite its effectiveness, visual servoing critically depends on the availability of reliable visual observations. In real-world manipulation, however, this requirement frequently fails to hold. During grasping, alignment, or object approach, the manipulator itself may partially or completely occlude the target. Meanwhile, fixed sensor placement can easily create unfavorable viewing angles or depth-dominant observations. Once the target or the end-effector leaves the camera view, the visual feedback loop is broken, and the manipulation process may fail. Ensuring stable observability of task-relevant features is therefore a central challenge for robust visual servoing~\cite{khoshelham2012accuracy,zhang2023occlusion}.

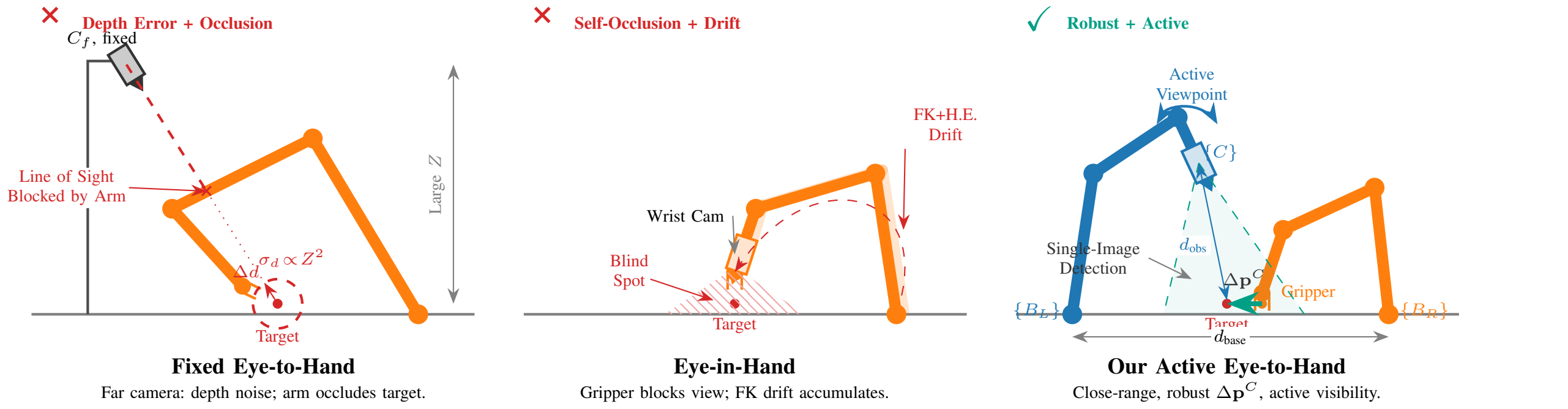
\begin{figure*}[t]
\centering
\begin{tikzpicture}[
    scale=0.76, transform shape,
    font=\small,
    >=Stealth,
    link/.style={line width=3.5pt, line cap=round},
    jointOurs/.style={circle, fill=clrOurs, inner sep=0pt, minimum size=8pt},
    jointSota/.style={circle, fill=clrSotaA, inner sep=0pt, minimum size=8pt},
    target/.style={circle, fill=clrRef, inner sep=0pt, minimum size=4pt},
    gript/.style={circle, fill=clrSotaA, inner sep=0pt, minimum size=4pt},
    pics/camOurs/.style={
        code={
            \fill[clrOurs!20, draw=clrOurs, thick] (-0.25,-0.15) rectangle (0.25,0.15);
            \fill[clrOurs] (0.25, -0.12) -- (0.45, 0) -- (0.25, 0.12) -- cycle;
        }
    },
    pics/camSota/.style={
        code={
            \fill[clrSotaA!20, draw=clrSotaA, thick] (-0.25,-0.15) rectangle (0.25,0.15);
            \fill[clrSotaA] (0.25, -0.12) -- (0.45, 0) -- (0.25, 0.12) -- cycle;
        }
    },
    pics/camFixed/.style={
        code={
            \fill[black!20, draw=black!80, thick] (-0.25,-0.15) rectangle (0.25,0.15);
            \fill[black!80] (0.25, -0.12) -- (0.45, 0) -- (0.25, 0.12) -- cycle;
        }
    }
]

\begin{scope}[xshift=0cm]
    \draw[thick, gray] (-0.5,0) -- (5.8,0);

    \node[target] (T1) at (3.0, 0.15) {};
    \node[font=\scriptsize, clrRef, below=5pt] at (T1.south) {Target};

    \draw[thick, black!70] (0.3,0) -- (0.3,3.6) -- (0.8,3.6);
    \coordinate (Cam1) at (0.85,3.55);
    \pic[rotate=-58] at (Cam1) {camFixed};
    \node[font=\scriptsize, black] at (0.5,3.9) {$C_f$, fixed};

    \draw[link, clrSotaA] (5.0,0) -- (3.5,2.5) coordinate (J1a);
    \draw[link, clrSotaA] (J1a) -- (1.5,1.5) coordinate (J2a);
    \draw[link, clrSotaA] (J2a) -- (2.5,0.4) coordinate (EE1);
    \fill[clrSotaA] (EE1) circle (0.12);
    \draw[thick, clrSotaA] (2.54,0.47) -- (2.76,0.36);
    \draw[thick, clrSotaA] (2.46,0.33) -- (2.69,0.22);
    \node[jointSota] at (5.0,0) {};
    \node[jointSota] at (J1a) {};
    \node[jointSota] at (J2a) {};

    \coordinate (Occ) at (1.99, 1.75);
    \draw[dashed, clrRef, thick] (Cam1) -- (Occ);
    \node[clrRef, font=\small\bfseries] at (Occ) {\texttimes};
    \draw[dotted, clrRef, thin] (Occ) -- (T1.center);

    \node[font=\scriptsize, clrRef, align=center, fill=white, inner sep=1pt]
        (LoSlabel) at (0.0, 1.8) {Line of Sight\\Blocked by Arm};
    \draw[->, clrRef, thin] (LoSlabel.east) -- (Occ);

    \draw[<->, gray, thin] (5.5,0.15) -- (5.5,3.55)
        node[midway, right, font=\scriptsize, gray, rotate=90, anchor=south] {Large $Z$};

    \draw[clrRef, dashed, thick] (T1.center) circle (0.35);
    \draw[->, clrRef, thin] (T1.center) -- (2.81, 0.45)
        node[above left=-2pt, font=\scriptsize, clrRef] {$\Delta d$};
    \node[font=\scriptsize, clrRef] at (3.2, 0.8) {$\sigma_d \!\propto\! Z^2$};

    \node[text=clrRef, anchor=north west] at (-0.5,4.5) {\textbf{\Large \texttimes}};
    \node[text=clrRef, anchor=north west, font=\bfseries\scriptsize] at (0.1,4.35) {Depth Error + Occlusion};

    \node[align=center, anchor=north] at (2.8,-0.5) {
        \textbf{Fixed Eye-to-Hand} \\
        {\scriptsize Far camera: depth noise; arm occludes target.}
    };
\end{scope}

\begin{scope}[xshift=7.0cm]
    \draw[thick, gray] (-0.5,0) -- (5.8,0);

    \node[target] (T2) at (2.5,0.15) {};
    \node[font=\scriptsize, clrRef, below] at (T2.south) {Target};

    \begin{scope}[opacity=0.2]
        \draw[link, clrSotaA] (4.9,0.05) -- (4.6,2.05);
        \draw[link, clrSotaA] (4.6,2.05) -- (2.9,1.55);
        \draw[link, clrSotaA] (2.9,1.55) -- (2.65,0.7);
    \end{scope}

    \draw[link, clrSotaA] (4.8,0) -- (4.5,2.0) coordinate (J2b1);
    \draw[link, clrSotaA] (J2b1) -- (2.8,1.5) coordinate (J2b2);
    \draw[link, clrSotaA] (J2b2) -- (2.5,0.6) coordinate (EE2);

    \fill[clrSotaA] (EE2) circle (0.12);
    \draw[thick, clrSotaA] (2.4,0.6) -- (2.4,0.35);
    \draw[thick, clrSotaA] (2.6,0.6) -- (2.6,0.35);

    \coordinate (Cam2) at (2.6,0.85);
    \pic[rotate=-108] at (Cam2) {camSota};
    \node[font=\scriptsize, black] (WCLabel) at (1.8,1.4) {Wrist Cam};
    \draw[->, gray, thin] (WCLabel.east) -- (2.5,0.85);

    \fill[pattern=north west lines, pattern color=clrRef!40]
        (2.4,0.6) -- (1.5,0) -- (3.5,0) -- (2.6,0.6) -- cycle;
    \node[font=\scriptsize, clrRef, align=center] at (1.0,0.6) {Blind\\Spot};
    \draw[->, clrRef, thin] (1.3,0.5) -- (2.2,0.2);

    \draw[->, clrRef, thin, dashed] (4.8,0) .. controls (5.4,1.8) and (3.6,2.2) .. (EE2);
    \node[font=\scriptsize, clrRef, align=center]
        (driftlabel) at (5.5,2.7) {FK+H.E.\\Drift};
    \draw[->, clrRef, thin] (driftlabel.south west) -- (4.9,1.2);

    \node[jointSota] at (4.8,0) {};
    \node[jointSota] at (J2b1) {};
    \node[jointSota] at (J2b2) {};

    \node[text=clrRef, anchor=north west] at (-0.5,4.5) {\textbf{\Large \texttimes}};
    \node[text=clrRef, anchor=north west, font=\bfseries\scriptsize] at (0.1,4.35) {Self-Occlusion + Drift};

    \node[align=center, anchor=north] at (2.5,-0.5) {
        \textbf{Eye-in-Hand} \\
        {\scriptsize Gripper blocks view; FK drift accumulates.}
    };
\end{scope}

\begin{scope}[xshift=14.0cm]
    \draw[thick, gray] (-0.5,0) -- (5.8,0);

    \node[target] (T3) at (2.5,0.15) {};
    \node[font=\scriptsize, clrRef, below] at (T3.south) {Target};

    \draw[link, clrSotaA] (4.8,0) -- (4.6,1.8) coordinate (J4c1);
    \draw[link, clrSotaA] (J4c1) -- (3.3,1.2) coordinate (J4c2);
    \draw[link, clrSotaA] (J4c2) -- (3.0,0.3) coordinate (EE4);
    \fill[clrSotaA] (EE4) circle (0.12);
    \draw[thick, clrSotaA] (2.9,0.3) -- (2.9,0.05);
    \draw[thick, clrSotaA] (3.1,0.3) -- (3.1,0.05);
    \node[jointSota] at (4.8,0) {};
    \node[jointSota] at (J4c1) {};
    \node[jointSota] at (J4c2) {};

    \node[gript] (GP) at (3.0,0.15) {};
    \node[font=\scriptsize, clrSotaA, right] at (3.15,0.3) {Gripper};

    \draw[link, clrOurs] (0.3,0) -- (0.6,2.0) coordinate (J3c1);
    \draw[link, clrOurs] (J3c1) -- (1.8,2.8) coordinate (J3c2);
    \draw[link, clrOurs] (J3c2) -- (2.1,2.15) coordinate (EE3);
    \coordinate (Cam3) at (2.1,2.15);
    \pic[rotate=-65] at (Cam3) {camOurs};
    \node[jointOurs] at (0.3,0) {};
    \node[jointOurs] at (J3c1) {};
    \node[jointOurs] at (J3c2) {};

    \fill[clrSotaB, opacity=0.06] (Cam3) -- (1.6,0) -- (3.6,0) -- cycle;
    \draw[dashed, clrSotaB, thin] (Cam3) -- (1.6,0);
    \draw[dashed, clrSotaB, thin] (Cam3) -- (3.6,0);

    \draw[->, clrSotaB, very thick, line width=1.5pt] (GP.center) -- (T3.center)
        node[midway, above=2pt, font=\scriptsize\bfseries, black!80] {$\Delta\mathbf{p}^C$};

    \draw[<->, clrOurs, thin] (Cam3) -- (2.5,0.2)
        node[pos=0.6, left=2pt, font=\scriptsize, clrOurs, fill=white, inner sep=1pt] {$d_{\text{obs}}$};

    \draw[<->, thick, clrOurs] (1.5,2.7) arc (150:30:0.5);
    \node[font=\scriptsize, clrOurs, align=center] at (2.0,3.25) {Active\\Viewpoint};

    \node[font=\scriptsize, black!80, align=center] (SFlabel) at (0.6,0.8) {Single-Image\\Detection};
    \draw[->, black!50, thin] (SFlabel.east) -- (2.0,0.5);

    \node[text=clrSotaB, anchor=north west] at (-0.5,4.5) {\textbf{\Large \checkmark}};
    \node[text=clrSotaB, anchor=north west, font=\bfseries\scriptsize] at (0.1,4.35) {Robust + Active};

    \node[font=\scriptsize, clrOurs, left=1pt] at (0.3,0.05) {$\{B_L\}$};
    \node[font=\scriptsize, clrSotaA, right=1pt] at (4.8,0.05) {$\{B_R\}$};
    \node[font=\scriptsize, clrOurs, above right=-3pt and -2pt] at (2.1,2.15) {$\{C\}$};
    \draw[<->, gray, thin] (0.3,-0.32) -- (4.8,-0.32)
        node[midway, fill=white, inner sep=1pt, font=\scriptsize, text=black] {$d_{\text{base}}$};

    \node[align=center, anchor=north] at (2.5,-0.5) {
        \textbf{Our Active Eye-to-Hand} \\
        {\scriptsize Close-range, robust $\Delta\mathbf{p}^C$, active visibility.}
    };
\end{scope}

\end{tikzpicture}
\caption{Motivation for active eye-to-hand visual servoing. From left to right, a fixed workspace camera observes from far away, so depth uncertainty and arm-induced occlusion accumulate during approach. Mounting the camera on the worker wrist restores proximity but introduces self-occlusion and calibration-error sensitivity near contact. Our asymmetric setup keeps an observer camera close to the action while preserving a shared view of both target and gripper, enabling relative servo cues under active viewpoint adjustment.}
\label{fig:motivation}
\end{figure*}

Most existing visual servoing methods implicitly assume that observation conditions remain sufficiently reliable throughout execution. In typical setups, cameras are mounted either in a fixed eye-to-hand configuration or in an eye-in-hand configuration rigidly attached to the manipulator. While these arrangements simplify system integration, they also constrain the available viewpoints and make the perception process vulnerable to self-occlusion and visibility loss, as illustrated in Fig.~\ref{fig:motivation}. Classical position-based visual servoing (PBVS) and image-based visual servoing (IBVS) pipelines~\cite{chaumette2006visual,chaumette2007visual} are effective when task-relevant features remain observable, but visibility recovery is usually outside the controller design. Learned or keypoint-based variants~\cite{bateux2018training,luo2023keypoint} strengthen feature extraction or pose estimation, yet they still largely rely on the assumption that the necessary observations remain available.

Several research directions attempt to mitigate this limitation, but each addresses only part of the problem. Occlusion-aware visual-servo constraints~\cite{zhang2023occlusion} improve robustness by enforcing visibility conditions, but the sensing geometry is still externally specified rather than actively maintained online. Active vision and next-best-view methods~\cite{newbury2022visibility,breyer2022nbv,li2024actpermoma} explicitly move the camera to improve visibility, yet they typically operate at planning frequencies and optimize scene coverage or information gain, instead of sustaining servo-rate visual feedback during manipulation. Dual-arm manipulation systems~\cite{ren2024dualarm,zhang2025ibvs} improve coordination and dexterity, but the visual feedback still comes from a static viewpoint that does not adapt to the evolving manipulation geometry.

Another limitation lies in how the servo signal itself is formed and whether the perception stack requires task-specific supervision. In standard position-based pipelines, the target and the end-effector are often estimated through separate kinematic and camera transformation chains before the control input is computed~\cite{chaumette2007visual}. This design makes the servo error sensitive to calibration inaccuracies, especially when the two estimates are produced independently. Meanwhile, recent learning-based active manipulation systems, including recent bimanual active-vision systems~\cite{av-aloha2024}, reduce hand-crafted engineering but often require task-specific teleoperation data, policy training, or task-level adaptation.

These gaps motivate a different formulation: rather than treating perception as a passive upstream module, we make observation an explicit and adaptive part of the servoing process. To this end, we propose an \emph{active observation framework} for robust visual servoing in robotic manipulation. In this framework, one arm of an existing dual-arm platform carries an RGB-D camera as an active observer and regulates the camera viewpoint to maintain or recover a shared target--gripper view when feasible. Specifically, the framework combines three key mechanisms: \emph{viewpoint adaptation} to reposition the observer arm toward more informative views, \emph{occlusion-triggered exploration} to restore a shared view when detection quality drops, and a \emph{Relative Servo Error} computed from the target--gripper displacement in a single camera image, avoiding differencing two independently transformed target and gripper estimates. By coupling active visibility maintenance with Relative Servo Error feedback from a single camera image, the proposed system uses recovered observations to update the servoing signal after occlusion begins.
Overall, the main contributions of this work are summarized as follows:

\begin{itemize}
\item We propose an active observation framework for robotic manipulation that repurposes one arm of an existing dual-arm platform as an active observer carrying an RGB-D camera. The observer maintains or recovers target--gripper visibility within a feasible shared-view workspace via anisotropy-aware two-phase positioning and occlusion-triggered exploration.
\item We formulate a Relative Servo Error from the target--gripper displacement observed in a single camera image, so that the two points are differenced in the shared camera frame before mapping to the robot base.
\item We validate the proposed framework on a real dual 6-DoF robot platform over \NumTrialsPerCond{} trials per object. Combining active observation with Relative Servo Error feedback from a single camera image achieves \OursOverallSuccess{} grasp success, compared with \FixedOverallSuccess{} for a fixed camera and \ReactiveSuccess{} for reactive viewpoint adjustment. The framework requires no additional active-vision arm, task-specific demonstrations, or task-specific fine-tuning.
\end{itemize}

\section{Methods}

\subsection{Overview}
\label{sec:overview}

\begin{figure}[t]
\centering
\resizebox{0.70\columnwidth}{!}{%
\begin{tikzpicture}[
    scale=0.60, transform shape,
    font=\normalsize,
    block/.style={
        rectangle,
        draw=black!65,
        line width=0.95pt,
        rounded corners=2.5pt,
        align=center,
        fill=white,
        minimum height=0.88cm,
        minimum width=2.85cm,
        inner sep=4pt
    },
    observerblock/.style={
        rectangle,
        draw=clrOurs!85!black,
        line width=1.15pt,
        rounded corners=2.5pt,
        align=center,
        fill=clrOurs!5,
        minimum height=0.9cm,
        minimum width=2.8cm,
        inner sep=4pt
    },
    workerblock/.style={
        rectangle,
        draw=clrSotaA!85!black,
        line width=1.15pt,
        rounded corners=2.5pt,
        align=center,
        fill=clrSotaA!8,
        minimum height=0.9cm,
        minimum width=2.8cm,
        inner sep=4pt
    },
    emphasis/.style={
        rectangle,
        draw=clrOurs!85!black,
        line width=1.35pt,
        rounded corners=2.5pt,
        align=center,
        fill=clrOurs!6,
        minimum height=0.95cm,
        minimum width=3.2cm,
        inner sep=5pt
    },
    decision/.style={
        diamond,
        draw=black!70,
        line width=0.95pt,
        aspect=2.5,
        align=center,
        fill=white,
        inner sep=0pt,
        font=\small\bfseries,
        minimum width=2.5cm
    },
    gate/.style={
        rectangle,
        draw=clrRef!75!black,
        line width=1.0pt,
        rounded corners=2pt,
        align=center,
        fill=white,
        minimum width=2.35cm,
        minimum height=0.66cm,
        inner sep=3pt,
        font=\scriptsize\bfseries,
        text=clrRef!85!black
    },
    arr/.style={-{Stealth[length=2.0mm, width=1.5mm]}, line width=1.0pt},
    auxarr/.style={-{Stealth[length=1.8mm, width=1.3mm]}, line width=1.0pt, color=black!70},
    bus/.style={line width=1.0pt, color=black!75},
    pausearr/.style={-{Stealth[length=1.8mm, width=1.3mm]}, dashed, line width=1.0pt, color=clrRef!85},
    softlabel/.style={font=\small\bfseries, text=black!62, inner sep=1.0pt},
    buslabel/.style={font=\small\bfseries, text=black!68, inner sep=1.0pt}
]

    \begin{scope}[shift={(0,0)}]
        \begin{scope}[yshift=0.56cm]

        \node[block, minimum width=2.3cm] (rgbd) at (-4.25,4.75) {RGB-D\\Camera};
        \node[observerblock] (det) at (-0.15,4.75) {Open-Vocab\\Detector};
        \node[block, minimum width=2.5cm, draw=clrOurs!65!black, fill=clrOurs!2] (trk) at (4.35,4.75) {CSRT\\Tracker};

        \coordinate (rgbbusL) at ($(rgbd.north)+(0,0.58)$);
        \coordinate (detbus) at ($(det.north)+(0,0.58)$);
        \coordinate (trkbus) at ($(trk.north)+(0,0.58)$);
        \draw[arr] (rgbd.north) -- (rgbbusL);
        \draw[bus] (rgbbusL) -- (trkbus);
        \node[buslabel] at ($(rgbbusL)!0.55!(trkbus)+(0,0.32)$) {RGB-D frames};
        \fill[black!75] (detbus) circle (0.045);
        \fill[black!75] (trkbus) circle (0.045);
        \draw[auxarr] (detbus) -- (det.north);
        \draw[auxarr] (trkbus) -- (trk.north);
        \draw[arr] (det.east) -- node[above, softlabel] {init / re-init} (trk.west);

        \node[block, draw=black!60, fill=gray!5, minimum width=7.0cm] (perc) at (0.0,2.75)
            {Perception: $({}^{C}\mathbf{p}_T, c_T),\; ({}^{C}\mathbf{p}_G, c_G)$};
        \coordinate (perclinkL) at ($(perc.north)+(-1.7,0)$);
        \coordinate (perclinkR) at ($(perc.north)+(1.7,0)$);
        \coordinate (detdrop) at ($(det.south)+(0,-0.48)$);
        \coordinate (trkdrop) at ($(trk.south)+(0,-0.48)$);
        \draw[arr] (det.south) -- (detdrop) -- (detdrop -| perclinkL) -- (perclinkL);
        \draw[arr] (trk.south) -- (trkdrop) -- (trkdrop -| perclinkR) -- (perclinkR);

        \node[decision] (dec) at (0.0,1.3) {Occluded?};
        \draw[arr] (perc.south) -- (dec.north);

        \node[observerblock] (expl) at (-4.2,-0.35) {\textbf{Explorer}\\{\footnotesize Recovery}};
        \node[observerblock] (foll) at (-4.2,-2.05) {\textbf{Follower}\\{\footnotesize Two-Phase Init}};
        \node[observerblock, fill=clrOurs!10] (obsctrl) at (-4.2,-3.75) {Observer Ctrl};

        \coordinate (noelbow) at (0.0,-2.05);
        \draw[arr] (dec.west) -| node[pos=0.17, above, softlabel] {yes} (expl.north);
        \draw[line width=1.0pt] (dec.south) -- (noelbow);
        \draw[arr] (noelbow) -- node[pos=0.62, above, softlabel] {no} (foll.east);
        \draw[arr] (expl) -- (foll);
        \draw[arr] (foll) -- (obsctrl);

        \node[font=\small\bfseries, color=clrOurs] (jl) at (-4.2,-5.05) {$\{B_L\}$ Joints};
        \draw[arr, color=clrOurs!85!black] (obsctrl) -- (jl);

        \node[emphasis] (relpose) at (4.35,-0.55) {\textbf{Relative Cue}\\[2pt] $\Delta\mathbf{p}^{C} = {^{C}\mathbf{p}_T} - {^{C}\mathbf{p}_G}$\\[2pt] $\mathbf{e}_p = {^{B_L}\mathbf{R}_C}\Delta\mathbf{p}^{C}$};
        \node[gate] (pausegate) at (4.35,-2.45) {Uncertainty\\Handling};
        \node[workerblock, fill=clrSotaA!12] (pbvs) at (4.35,-3.85) {Worker PBVS};
        \node[font=\small\bfseries, color=clrSotaA] (jr) at (4.35,-5.05) {$\{B_R\}$ Joints};

        \coordinate (statejoin) at (relpose.north |- perc.east);
        \draw[auxarr] (perc.east) -- node[pos=0.5, above, softlabel] {state} (statejoin) -- (relpose.north);
        \draw[arr] (relpose) -- node[pos=0.48, right, softlabel] {$\mathbf{e}_p$} (pausegate);
        \draw[arr] (pausegate) -- node[pos=0.52, right, softlabel] {$g(t)\mathbf{e}_p$} (pbvs);
        \draw[arr, color=clrSotaA!85!black] (pbvs) -- (jr);

        \coordinate (pauselead) at ($(expl.east)+(0.18,0)$);
        \coordinate (pauseacross) at (0.55,-0.35);
        \coordinate (pausedown) at (pauseacross |- pausegate.center);
        \coordinate (pauseout) at ($(pausegate.west)+(-0.28,0)$);
        \draw[color=clrRef!85, line width=1.0pt] (expl.east) -- (pauselead);
        \draw[pausearr] (pauselead) -- (pauseacross) -- (pausedown) -- node[pos=0.55, above, softlabel, text=clrRef!85!black] {pause / cooldown} (pauseout) -- (pausegate.west);
        \end{scope}
    \end{scope}

\end{tikzpicture}%
}
\caption{Framework overview. The shared RGB-D stream feeds detection and tracking; the fused target/gripper state yields the camera-frame cue $\Delta\mathbf{p}^{C}$ and the rotated servo error $\mathbf{e}_p$. On the observer side, the follower performs two-phase initialization while the explorer handles occlusion recovery; on the worker side, confidence-based uncertainty handling gates PBVS.}
\label{fig:framework}
\end{figure}
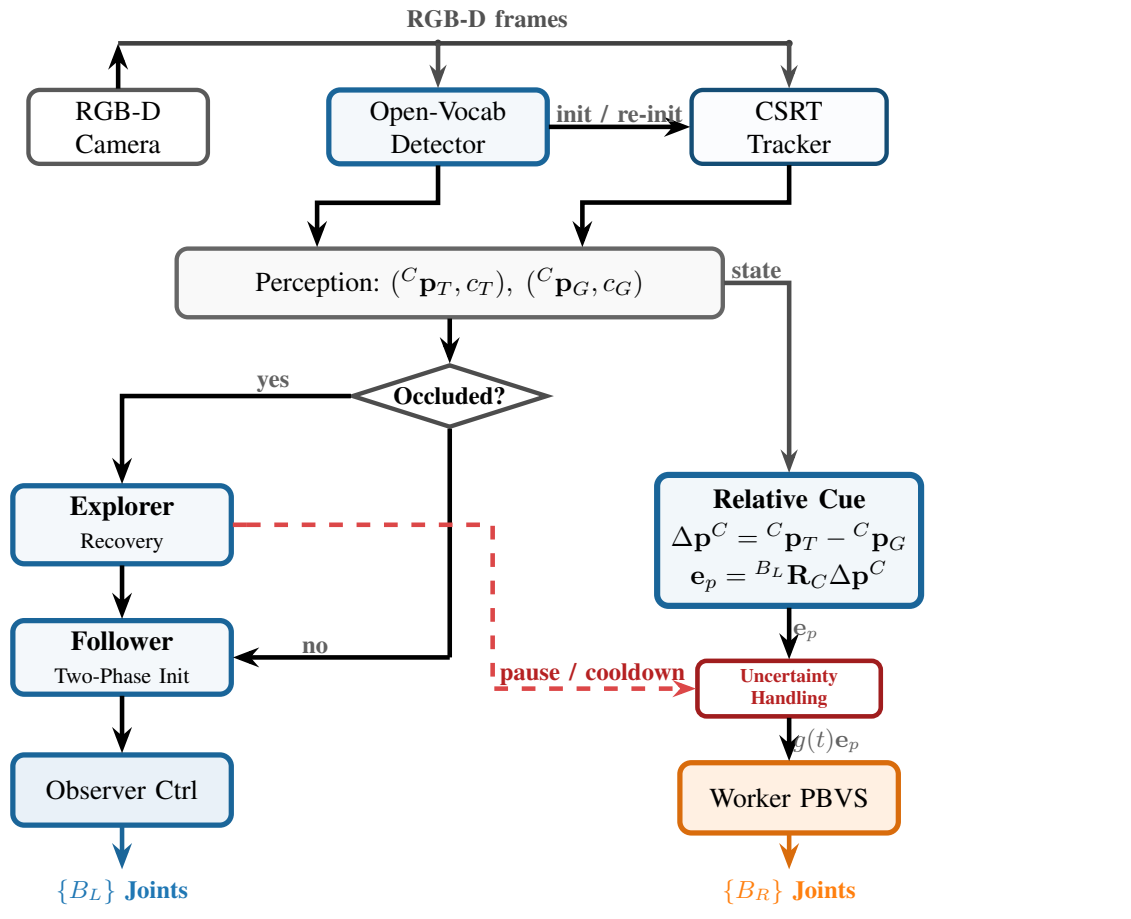

To address visibility loss and calibration-sensitive feedback, we propose an active observation framework for robotic grasping. The framework uses one arm of an existing dual-arm platform as an \emph{observer arm} carrying an RGB-D camera and the other as a \emph{worker arm} for grasping. The observer repositions to maintain or recover a shared target--gripper view when feasible, while the worker uses the resulting visual feedback for servo execution.
The method couples three components: a \emph{Relative Servo Error} robust to translational calibration error, observer-side viewpoint regulation with occlusion-triggered recovery, and a confidence-aware worker servo law with inter-arm safety monitoring (Fig.~\ref{fig:framework}).

\subsection{Relative Servo Error Computation}
\label{sec:same_frame}

We begin by defining the Relative Servo Error, because both the observer's viewpoint policy and the worker's grasping law are built around this shared target--gripper signal rather than two independently transformed poses. This design reduces the influence of translational hand--eye calibration error and yields a compact visual cue for downstream control.
Using the geometry in Fig.~\ref{fig:motivation} (right), we use three coordinate frames: the observer-arm base $\{B_L\}$, the worker-arm base $\{B_R\}$, and the camera frame $\{C\}$. Here, $^{A}\mathbf{R}_{B}$ and $^{A}\mathbf{t}_{B}$ denote the rotation and translation of frame $\{B\}$ expressed in frame $\{A\}$. The two robot bases are mounted on the same planar platform with a fixed lateral offset $d_{\text{base}}$ and approximately identical orientation.

From the shared RGB-D observation, the observer detects both the target and the gripper tip, and back-projects their image centers to 3D points $^{C}\mathbf{p}_T$ and $^{C}\mathbf{p}_G$ in the camera frame. Their relative displacement is then written as
\begin{equation}
\Delta\mathbf{p}^{C}=\,^{C}\mathbf{p}_T-\,^{C}\mathbf{p}_G.
\label{eq:relative_disp}
\end{equation}

Because both points are measured in the same camera frame, this displacement is computed before either point is transformed into a robot base frame.
To generate the servo correction for the worker arm, we rotate this relative displacement into the base frame of the observer arm:
\begin{equation}
\mathbf{e}_p =\, ^{B_L}\mathbf{R}_C\,\Delta\mathbf{p}^{C}.
\label{eq:same_frame_error}
\end{equation}
Because the two bases are mounted with approximately identical orientation, $^{B_R}\mathbf{R}_{B_L}\approx \mathbf{I}$, this vector is used directly as the worker-frame Cartesian correction up to a small base-alignment error: writing $^{B_R}\mathbf{R}_{B_L}=\mathbf{I}+\delta\mathbf{R}_b$ adds a bias of order $O(\|\delta\mathbf{R}_b\|\,\|\Delta\mathbf{p}^{C}\|)$, distinct from the camera rotational calibration term discussed below. This representation suppresses the effect of translational hand--eye error at the level of the Relative Servo Error, while preserving the geometric direction needed for servoing.

Residual rotational calibration error $\delta\mathbf{R}$ biases $\mathbf{e}_p$ by $O(\|\delta\mathbf{R}\|\,\|\Delta\mathbf{p}^C\|)$, and this bias shrinks as the gripper approaches the target. This error signal drives both the active viewpoint controller, which continuously updates the detections used to compute $\Delta\mathbf{p}^C$, and the worker arm's confidence-gated grasping law.

\subsection{Active Viewpoint Control}
\label{sec:active_observation}

This servo error is only as accurate as the underlying detections of $^{C}\mathbf{p}_T$ and $^{C}\mathbf{p}_G$, so the observer arm actively selects viewpoints that improve their reliability. During normal approach, the observer aligns high-precision sensor axes with the most critical error components. When occlusion disrupts detection, it searches for a viewpoint that restores visibility.

\noindent\textbf{Perception.}
An open-vocabulary detector localizes both the target and the gripper tip from natural-language prompts. Between detector updates, a lightweight visual tracker maintains bounding boxes at the camera frame rate. A temporal filter rejects single-step outliers and smooths accepted positions with a first-order low-pass filter. Each detected center is back-projected to 3D using the camera intrinsics and the median valid depth within the detected region, yielding the positions $^{C}\mathbf{p}_T$ and $^{C}\mathbf{p}_G$. All numerical constants used throughout this section are fixed across experiments. Key runtime settings appear in Table~\ref{tab:setup}.

\noindent\textbf{Two-phase viewpoint initialization.}
To improve the precision of this Relative Servo Error, we exploit the anisotropic noise structure of RGB-D sensors. At close range, the lateral uncertainty $\sigma_\ell$ is lower than the depth-axis uncertainty $\sigma_d$, since $\sigma_\ell \propto Z$ whereas $\sigma_d \propto Z^2$~\cite{khoshelham2012accuracy}. We therefore position the observer arm so that high-precision lateral axes align with the error components that matter most.

Let $\mathbf{d}_f(\theta_{J5}) = [\cos\theta_{J5},\,\sin\theta_{J5},\,0]^\top$ denote the worker arm's finger direction derived from the wrist yaw, and $\mathbf{d}_\perp = \hat{\mathbf{z}} \times \mathbf{d}_f$ the perpendicular direction, where $\hat{\mathbf{z}}=[0,0,1]^\top$ is the vertical unit vector. Both directions are expressed in the base frame. We decompose this servo error along these task-relevant axes:
\begin{equation}
\mathbf{e}_p = e_f\,\mathbf{d}_f + e_\perp\,\mathbf{d}_\perp + e_z\,\hat{\mathbf{z}},
\label{eq:error_decomp}
\end{equation}
where $e_f = \mathbf{d}_f^\top\mathbf{e}_p$, $e_\perp = \mathbf{d}_\perp^\top\mathbf{e}_p$, and $e_z = \hat{\mathbf{z}}^\top\mathbf{e}_p$. The observer arm positions itself at a standoff distance $d_{\text{obs}}$ from the midpoint of the two detected objects:
\begin{equation}
\mathbf{p}_{\text{obs}} = \tfrac{1}{2}\!\left({^{B_L}\!\mathbf{p}_T} + {^{B_L}\!\mathbf{p}_G}\right) + d_{\text{obs}}\,\mathbf{d}_{\text{view}},
\label{eq:obs_position}
\end{equation}
where ${^{B_L}\!\mathbf{p}_T}$ and ${^{B_L}\!\mathbf{p}_G}$ are the detected positions transformed to the left base frame, and $\mathbf{d}_{\text{view}}$ selects the viewing direction. The \emph{side view} sets $\mathbf{d}_{\text{view}} = \mathbf{d}_\perp$, so that $e_f$ and $e_z$ are measured primarily along the lower-noise lateral axes. The \emph{front view} sets $\mathbf{d}_{\text{view}} = \mathbf{d}_f$, so the lateral axes capture $e_\perp$ and $e_z$, while the vertical component is reinforced in both views. This two-phase cycle runs once at the start of each approach episode. Afterward, standard PBVS resumes with the full 3D Relative Servo Error. Later visibility losses are handled by exploration rather than by repeating the cycle.

\noindent\textbf{Occlusion-triggered exploration.}
If detection confidence drops below a threshold for consecutive cycles, the system pauses the worker arm and samples candidate viewpoints at azimuth increments around the last known focus point. Each candidate at azimuth $\phi$ is scored by a weighted sum of four detection-quality cues:
\begin{equation}
Q(\phi) = w_c\,\bar{c}(\phi) + w_a\,\hat{a}(\phi) + w_d\,\hat{d}(\phi) + w_s\,\hat{s}(\phi),
\label{eq:quality_score}
\end{equation}
where $\bar{c}$ is the mean detection confidence, $\hat{a}$ the normalized bounding-box area, $\hat{d}$ the depth-validity ratio, and $\hat{s}$ the temporal stability, each in $[0,1]$. The weights are fixed across all experiments as $(w_c,w_a,w_d,w_s)=(0.4,0.2,0.2,0.2)$.

The viewpoint with the highest $Q$ is selected. If no sampled candidate satisfies the recovery criterion, coordinate descent refines the best-scoring candidate. Exploration terminates when both confidences recover for consecutive cycles, or after a timeout. If the timeout expires without recovery, both arms enter a safe halt. The requirement that both confidences recover for consecutive cycles doubles as a \emph{post-recovery cooldown}: the worker arm remains paused throughout to prevent premature re-engagement from transient detections. Once visibility is restored, the updated Relative Servo Error feeds the gated control law.

\subsection{Confidence-Aware Servo Grasping}
\label{sec:servo_grasping}

The worker arm approaches and grasps the target using the Relative Servo Error~\eqref{eq:same_frame_error} and the detection confidences $c_T$, $c_G$ provided by the active observer. To suppress motion during uncertain detections, we define a confidence gate:
\begin{equation}
g(t) = \mathbf{1}\!\left[\sqrt{c_T(t)\,c_G(t)} \ge c_{\text{gate}}\right],
\label{eq:confidence_gate}
\end{equation}
where $c_T$ and $c_G$ are the target and gripper detection confidences at cycle $t$. The geometric mean ensures that the gate closes whenever either confidence falls to zero and penalizes imbalanced values. The threshold $c_{\text{gate}}$ is fixed. The gated control law drives the worker arm:
\begin{equation}
^{B_R}\mathbf{v} = g(t)\,\lambda\,\mathbf{e}_p,
\label{eq:pbvs_control}
\end{equation}
with gain $\lambda=0.5\,\mathrm{s}^{-1}$ and Cartesian velocity saturation $\|^{B_R}\mathbf{v}\|\le 0.05\,\mathrm{m/s}$. We retain a pure proportional law because the target is stationary and measurements are low-pass filtered. The end-effector orientation is held vertical. The resulting 3-DOF Cartesian command is mapped to joint velocities via damped-least-squares resolved-rate control.

Grasp execution proceeds through three phases, each narrowing the active error components defined above.

\noindent\textbf{Approach.} The full 3D law drives the gripper toward the target, and the confidence gate pauses motion when confidence is low.

\noindent\textbf{Fine alignment.} Once the gripper reaches a proximity zone, the vertical component is locked and only lateral corrections are applied. A range gate verifies that the target lies within a capture window for $K_{\text{stable}}$ consecutive frames.

\noindent\textbf{Blind grasp.} When the gripper--target distance falls below $\epsilon_\text{g}$, lateral commands are frozen and only approach-axis motion continues until contact. The preceding fine-alignment gate bounds residual lateral error. Throughout all phases, the observer arm repositions according to the viewpoint rule above. If occlusion recurs, exploration is activated.

\noindent\textbf{Safety.}
\label{sec:safety}
A capsule-based monitor approximates each link as a capsule of radius $r_{\text{link}}$, freezes both arms when the minimum inter-capsule distance drops below a stop threshold, and scales velocity linearly at intermediate distances; the same freeze applies when exploration times out without confidence recovery. The collision monitor thus joins the Relative Servo Error, the active observer arm, and the confidence gate to close the control loop: the worker arm advances only when detection confidence is sufficient and the inter-arm proximity constraint permits motion.

\section{Experiments}
\label{sec:experiments}
\subsection{Experimental Setup}
\label{sec:setup}
\noindent\textbf{Hardware and protocol.}
We evaluate on a dual-arm Piper platform comprising two 6-DoF arms separated by 0.60\,m and an Intel RealSense D435i. All methods share the same perception front-end, namely GroundingDINO~\cite{liu2023grounding} with CSRT tracking~\cite{lukezic2017csrt}; the shared hardware and protocol settings appear in Table~\ref{tab:setup}. Each method is tested over \NumTrialsPerCond{} randomized placements per object from the nine-object pool listed in Table~\ref{tab:comparison}.

\begin{figure}[t]
\centering
\includegraphics[width=\columnwidth]{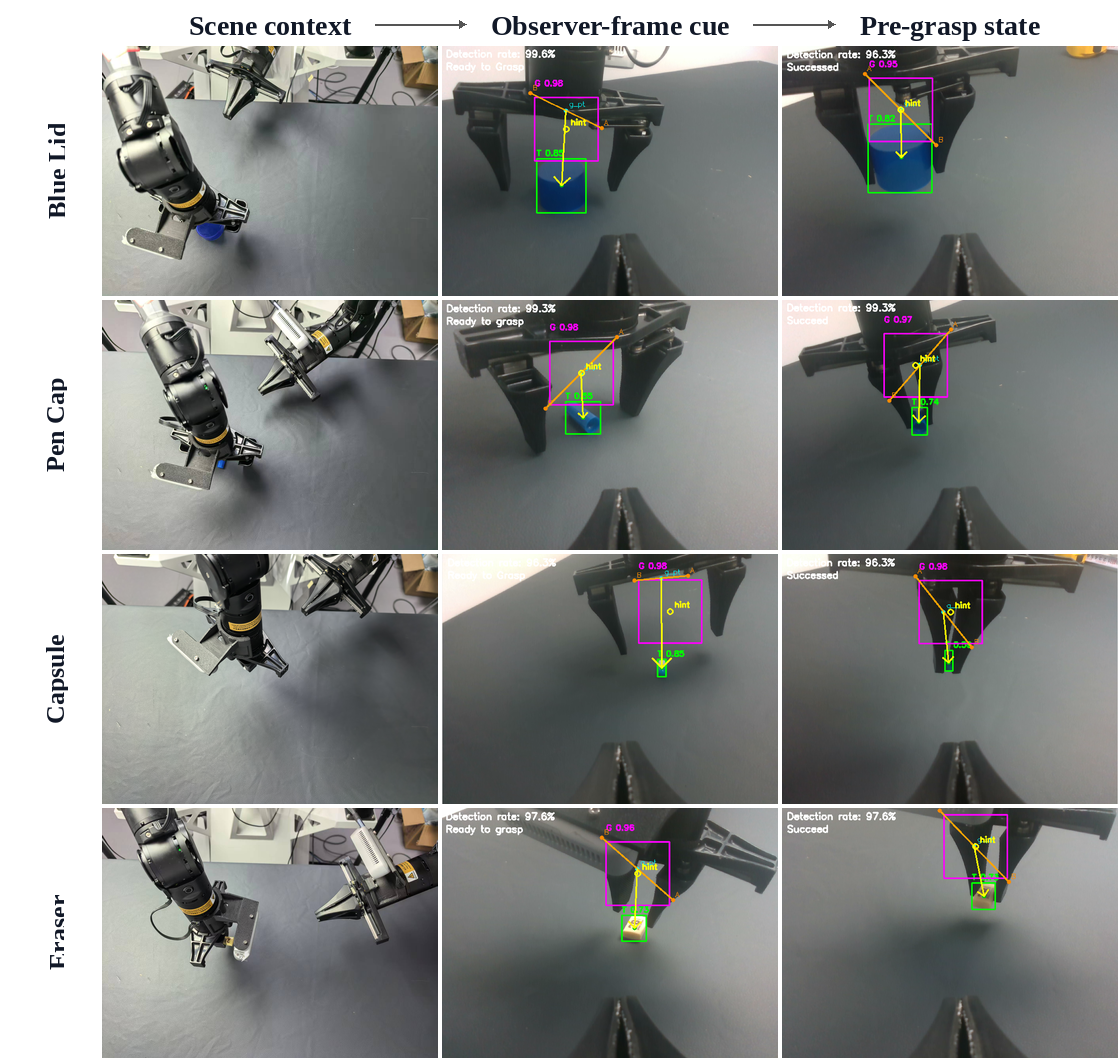}
\vspace{-2mm}
\caption{Observer-view visualization of the servo process for four representative objects: Blue Lid, Pen Cap, Capsule, and Eraser. Each row shows the scene, the relative displacement cue $\Delta\mathbf{p}^{C}$, and the aligned pre-grasp state.}\label{fig:object_gallery}
\vspace{-4mm}
\end{figure}

\begin{figure*}[t!]
\centering
\includegraphics[width=\textwidth]{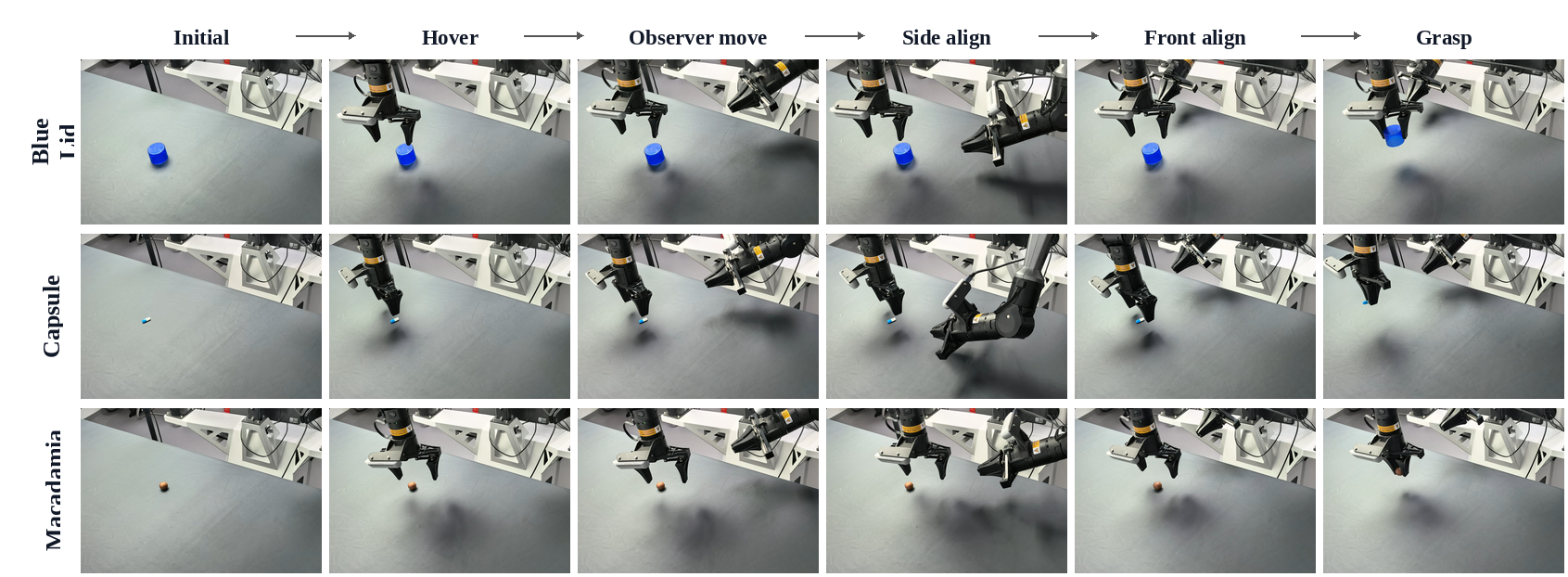}
\vspace{-2mm}
\caption{Representative workflow on three objects from the evaluation pool. Each row shows one trial from initial pose to grasp. The six columns correspond to the stages labeled in the figure: Initial, Hover, Observer move, Side align, Front align, and Grasp.}
\label{fig:workflow_sequence}
\vspace{-3mm}
\end{figure*}

\noindent\textbf{Task.}
Each trial (Fig.~\ref{fig:workflow_sequence}) requires the worker arm to approach, align, and grasp one object from the nine-object tabletop pool listed in Table~\ref{tab:objects}, under visual servoing guided by the observer. The nine objects range from centimeter-scale items that occupy only tens of pixels to palm-sized items with large image footprints, and include both regular and irregular geometries (Fig.~\ref{fig:object_gallery}).
\begin{table}[t]
\centering
\caption{\textbf{Shared hardware and protocol settings.}}
\label{tab:setup}
\footnotesize
\setlength{\tabcolsep}{3.5pt}
\renewcommand{\arraystretch}{1.08}
\begin{tabular}{@{}>{\centering\arraybackslash}m{0.20\columnwidth}>{\centering\arraybackslash}m{0.75\columnwidth}@{}}
\toprule[1.5pt]
Item & Setting \\
\midrule
Platform & Dual-arm Piper, 0.60\,m base offset, D435i camera \\
Perception & GroundingDINO + CSRT shared by all eye-to-hand methods \\
Control & PBVS at \ControlHz\,Hz, $\lambda{=}0.5$\,s$^{-1}$, $\|v\|{\le}0.05$\,m/s, confidence gate $c_{\text{gate}}{=}0.5$, observer standoff $d_{\text{obs}}{=}\ObsDistM$\,m, $K_{\text{stable}}{=}3$, blind-grasp threshold $\epsilon_\text{g}=\BlindGraspDist$\,m \\
Recovery & Trigger if confidence ${<}\ExplTriggerConf$ for \ExplTriggerN{} cycles; recover at ${\ge}\RecoveryConf$ for \RecoveryK{} cycles; timeout \ExplTimeoutS\,s \\
Safety & Capsule stop/slow thresholds \CapsuleStop/\CapsuleSlow\,m, link radius \LinkRadius\,m \\
\bottomrule[1.5pt]
\end{tabular}
\vspace{-2mm}
\end{table}
\noindent\textbf{Compared methods.}
The four eye-to-hand rows in Table~\ref{tab:comparison} share a common Relative Servo Error and the same PBVS law, with differences only in the observer arm's policy. Fixed keeps a workspace camera stationary, Static mounts the camera on the observer arm without online motion, Reactive triggers detours only after severe occlusion, and our method adds continuous following with exploration scored by observation quality. Eye-in-Hand mounts the camera on the worker arm's wrist and serves as an architectural wrist-camera reference rather than a controlled eye-to-hand comparison.

\noindent\textbf{Metrics.}
We report four metrics. Success rate is the fraction of trials in which the gripper closes within $\epsilon_\text{g}$ under closed-loop servoing. Position error is the $\ell_2$ distance between gripper and target at closure. Recovery time is measured from the severe-occlusion trigger to confidence recovery on triggered trials only. Cycle time is the mean wall-clock duration per control iteration.
\subsection{Main Results}
\label{sec:results}
Among the five methods evaluated in Table~\ref{tab:comparison}, the four eye-to-hand rows show a graded trend: success rises from \FixedOverallSuccess{} for Fixed to \StaticSuccess{} for Static, \ReactiveSuccess{} for Reactive, and \OursOverallSuccess{} for our method, while the closure error drops from \FixedPoseErr\,mm to \OursPoseErr\,mm. Because all four rows share the same Relative Servo Error and the same PBVS law, this trend is consistent with progressively more active observer policies. Eye-in-Hand scores \EIHSuccess{} success at \EIHPoseErr\,mm error; the gap is consistent with the visibility loss expected when the camera is attached to the manipulator.

The results across object categories help explain where the performance gains arise. For Blue Lid, Eraser, and Battery, which occupy large image areas, a major challenge is calibration drift near contact. Our method reaches 96--100\% on these objects because the Relative Servo Error formulation suppresses translational hand--eye offsets, so that the remaining sensitivity is mainly associated with the residual rotational calibration error analyzed in Section~\ref{sec:same_frame}, whose effect decreases as the gripper approaches the target.

The most discriminating test involves small objects: Hex Bolt, Bottle Cap, and Pen Cap. These items span only tens of pixels, and the fixed and static baselines cannot actively bring their viewpoint to the depth sensor's high-precision range, leading to low signal-to-noise ratio (SNR) and significant depth-axis jitter~\cite{khoshelham2012accuracy}. While these baselines degrade on small objects, our method maintains 88--92\% success: when a shared view is feasible, the active observer repositions to \ObsDistM\,m---within the sensor's high-precision close range---which raises the pixel-per-centimeter resolution and keeps the target within the sensor's low-noise lateral regime.

For irregular geometries like Capsule, Peanut, and Macadamia, the two-phase strategy appears particularly important. A single static viewpoint often yields a projected error that masks misalignments along the camera's depth axis. By verifying alignment from two orthogonal directions, our method suppresses these residual errors, reaching 92--96\% success on this challenging subset.

\noindent\textbf{Failure Mode Analysis.}
The remaining failures fall into three regimes: \emph{dynamic response lag}, where the worker arm outpaces observer repositioning and causes transient occlusion before the viewpoint adapts; \emph{workspace constraints}, where joint limits or self-collision avoidance block the ideal explorer-sampled viewpoint; and \emph{perception outliers}, where specular reflections from metallic objects (e.g., Hex Bolt) cause temporary depth holes beyond temporal-filter smoothing.
\begin{table*}[t]
\centering
\caption{\textbf{Per-object grasp performance across the nine-object pool.} Object cells report S/E/R\,=\,success\,(\%)/closure error\,(mm)/recovery\,(s), with recovery measured on triggered severe-occlusion trials. Avg.\ S/E aggregates the nine objects (\NumTrialsPerCond{} trials each); Avg.\ R/C reports recovery\,(s) and mean cycle time\,(ms). Bold marks the best success, error, and recovery; cycle time is reported as overhead.}
\label{tab:comparison}
\label{tab:objects}
\footnotesize
\setlength{\tabcolsep}{2.5pt}
\renewcommand{\arraystretch}{1.2}
\resizebox{\linewidth}{!}{
\begin{tabular}{@{}lccccccccc|cc@{}}
\toprule[1.5pt]
Method & Blue Lid & Eraser & Battery & Hex Bolt & Bttl.\ Cap & Pen Cap & Capsule & Peanut & Macad. & Avg.\ S/E & Avg.\ R/C \\
\midrule
Eye-in-Hand & 70/13.2/-- & 54/17.8/-- & 66/15.4/-- & 30/24.6/-- & 38/22.3/-- & 44/21.1/-- & 58/16.5/-- & 48/19.2/-- & 56/14.6/-- & \EIHSuccess/\EIHPoseErr & --/\EIHCycle \\
Fixed Obs. & 82/8.4/6.3 & 68/12.8/8.1 & 74/10.6/7.2 & 44/17.9/11.4 & 50/16.2/10.2 & 54/14.8/9.5 & 70/9.8/7.0 & 58/12.9/8.5 & 76/10.7/5.6 & \FixedOverallSuccess/\FixedPoseErr & \FixedRecovery/\FixedCycle \\
Static Obs. & 86/7.6/4.8 & 74/12.1/7.3 & 78/9.3/5.9 & 52/15.8/9.6 & 56/14.5/8.4 & 68/10.8/6.5 & 84/8.9/5.7 & 64/13.6/7.1 & 72/10.1/4.5 & \StaticSuccess/\StaticPoseErr & \StaticRecovery/\StaticCycle \\
Reactive & 90/6.5/3.2 & 82/10.9/5.1 & 86/8.1/3.8 & 60/14.2/6.7 & 64/12.8/5.8 & 66/11.5/4.4 & 80/9.6/3.9 & 74/11.4/4.9 & 80/8.6/3.5 & \ReactiveSuccess/\ReactivePoseErr & \ReactiveRecovery/\ReactiveCycle \\
\textbf{Ours} & \textbf{100/3.8/1.2} & \textbf{96/6.5/2.1} & \textbf{98/5.1/1.5} & \textbf{88/9.4/2.8} & \textbf{90/8.7/2.4} & \textbf{92/7.6/1.9} & \textbf{96/5.3/1.6} & \textbf{92/6.9/2.0} & \textbf{94/6.2/1.4} & \textbf{\OursOverallSuccess/\OursPoseErr} & \textbf{\OursRecoveryMeanS$\pm$\OursRecoveryStdS}/\OursCycle \\
\bottomrule[1.5pt]
\end{tabular}}
\par\vspace{1mm}
{\raggedright\footnotesize Note: S/E/R\,=\,success rate (\%)/closure error (mm)/recovery time (s); recovery is averaged over triggered severe-occlusion trials (e.g., $n_{\mathrm{trig}}{=}18$ for Ours, $31$ for Fixed). Avg.\ S/E is the mean over the nine objects (\NumTrialsPerCond{} trials each); Avg.\ R/C reports recovery (s)/mean control-cycle time (ms). ``--'' denotes Eye-in-Hand, which has no dual-detection recovery; Bttl.\ Cap $=$ Bottle Cap, Macad.\ $=$ Macadamia. Eye-in-Hand uses a wrist camera; the four eye-to-hand rows share perception/PBVS and differ only in observer policy.\par}
\vspace{-3mm}
\end{table*}

\subsection{Ablation Study}
\label{sec:ablation}
Table~\ref{tab:ablation} isolates each module's contribution by disabling one component at a time while keeping the rest of the pipeline fixed, reporting success and closure error as the endpoints most directly tied to grasp reliability.

\noindent\textbf{Relative Servo Error and Two-Phase.}
Removing the Relative Servo Error formulation (w/o RSE) forces the system to estimate target and gripper positions through independent kinematic chains, making the servo correction sensitive to the cumulative calibration error of both arms. This re-exposes the system to translational bias~\cite{chaumette2007visual}, causing the closure error to rise from \OursPoseErr\,mm to \AblSFPoseErr\,mm, the largest error increase across all four variants. This indicates that the differencing property of the Relative Servo Error contributes most to grasp precision among the variants tested. Removing the two-phase viewpoint schedule (w/o TP) eliminates the orthogonal verification step before the final approach. In this case, the observer arm can only verify alignment from a single direction, leaving perpendicular error components unchecked. Success drops to \AblTPSuccess{} and error rises to \AblTPPoseErr\,mm, indicating that pre-grasp alignment for non-spherical objects is sensitive to viewpoint coverage.

\noindent\textbf{Explorer and Uncertainty Handling.}
Without quality-scored exploration (w/o EX), triggered severe-occlusion events are not actively recoverable: the worker arm pauses and eventually times out. Disabling the explorer produces the largest success drop among all variants, from \OursOverallSuccess{} to \AblNoExplSuccess{}, which indicates that active recovery, rather than passive waiting, contributes most to the overall success rate under occlusion in the unstructured placements tested here. By contrast, disabling the confidence-aware uncertainty handling (w/o UH) results in a smaller success loss to \AblSMSuccess{} but a noticeable error increase to \AblSMPoseErr\,mm. This is because the worker arm continues to move even when detections are unstable, leading to intermittent jitter during the final alignment.
\begin{table}[t]
\centering
\caption{\textbf{Ablation study.} Each row disables one module from the full pipeline.}
\label{tab:ablation}
\footnotesize
\setlength{\tabcolsep}{3.5pt}
\renewcommand{\arraystretch}{1.08}
\resizebox{\columnwidth}{!}{%
\begin{tabular}{@{}lcccccc@{}}
\toprule[1.5pt]
Variant & RSE & TP & EX & UH & Succ.\ (\%) & Pos.\ Err.\ (mm) \\
\midrule
\textbf{Full} & \cmark & \cmark & \cmark & \cmark & \textbf{\OursOverallSuccess} & \textbf{\OursPoseErr} \\
w/o RSE & \xmark & \cmark & \cmark & \cmark & \AblSFSuccess & \AblSFPoseErr \\
w/o TP & \cmark & \xmark & \cmark & \cmark & \AblTPSuccess & \AblTPPoseErr \\
w/o EX & \cmark & \cmark & \xmark & \cmark & \AblNoExplSuccess & \AblNoExplPoseErr \\
w/o UH & \cmark & \cmark & \cmark & \xmark & \AblSMSuccess & \AblSMPoseErr \\
\bottomrule[1.5pt]
\end{tabular}}
\par\vspace{0.5mm}
{\raggedright\footnotesize Note: RSE = Relative Servo Error, TP = Two-Phase, EX = Explorer, UH = Uncertainty Handling.\par}
\vspace{-3mm}
\end{table}

\noindent\textbf{Calibration sensitivity.}
As a supplementary stress test rather than a standard component ablation, we inject random-direction offsets $\Delta\mathbf{t}$ of $\{10, 20, 30\}$\,mm into the translational component of the hand--eye calibration and compare the full system against w/o~RSE at each offset level. This test isolates the translational part of the hand--eye offset; residual rotational calibration error is not injected here and is covered by the bound in Section~\ref{sec:same_frame}. The full system degrades gradually, retaining \CalPertThirtyFullSucc{} success at the 30\,mm offset, whereas w/o~RSE deteriorates sharply to \CalPertThirtyNoSFSucc{}, consistent with the translational differencing of the relative formulation.

\section{Conclusion}
\label{sec:conclusion}

We present an active observation framework for visual servoing in which one arm of an existing dual-arm platform acts as an active observer and repositions to maintain or recover a shared target--gripper view when feasible. On a dual 6-DoF Piper platform, the framework achieves \OursOverallSuccess{} grasp success with \OursPoseErr{}\,mm closure error, versus \ReactiveSuccess{} for the reactive baseline and \FixedOverallSuccess{} for a fixed camera, without task-specific fine-tuning or demonstrations.
The method assumes stationary tabletop targets, two arm bases mounted on a common plane with approximately parallel orientations, and approximately vertical top-down grasps. Performance degrades under dynamic response lag, where the worker arm outpaces view recovery; workspace constraints, where collision or joint limits leave no admissible observer arm pose; and perception outliers under extreme lighting. Future work will address learning-based viewpoint prediction and full SE(3) relative servoing for non-parallel arm bases.

\bibliographystyle{IEEEtran}
\bibliography{references}

\end{document}